\documentclass[letterpaper, 10 pt, conference]{ieeeconf}  

\IEEEoverridecommandlockouts                              
\usepackage{comment}
\usepackage[utf8]{inputenc}
\usepackage[T1]{fontenc}

\usepackage{graphicx}
\usepackage{caption}
\usepackage{amsmath}
\usepackage{amssymb}
\usepackage{float}
\usepackage{threeparttable}
\usepackage{xcolor}
\usepackage[export]{adjustbox}
\usepackage{multirow}
\usepackage{algorithm}
\usepackage{algpseudocode}
\usepackage{colortbl}
\usepackage{dsfont}
\usepackage{comment}
\usepackage{hyperref}
\hypersetup{colorlinks=true,linkcolor=blue,citecolor=blue,urlcolor=black}
\usepackage{cite}
\usepackage{booktabs}
\usepackage{fancyhdr} 
\usepackage{tikz}

\newcommand\copyrighttext{%
  \footnotesize \textcopyright 2025 IEEE. Personal use of this material is permitted.  Permission from IEEE must be obtained for all other uses, in any current or future media, including reprinting/republishing this material for advertising or promotional purposes, creating new collective works, for resale or redistribution to servers or lists, or reuse of any copyrighted component of this work in other works.}
\newcommand\copyrightnotice{%
\begin{tikzpicture}[remember picture,overlay]
\node[anchor=south,yshift=10pt] at (current page.south) 
  {\fbox{\parbox{\dimexpr\textwidth-\fboxsep-\fboxrule\relax}{\copyrighttext}}};
\end{tikzpicture}%
}

\newcommand{\MSbl}[1]{{\color{black}#1}}
\begin{document}
\title{\LARGE \bf
Real-Time Adaptive Motion Planning via Point Cloud-Guided, Energy-Based Diffusion and Potential Fields}
\author{Wondmgezahu Teshome, Kian Behzad, Octavia Camps, Michael Everett, Milad Siami, and Mario Sznaier%
\thanks{\textcolor{black}{This work was 
supported by
NSF grants CNS--2038493 and CMMI--2208182, AFOSR grant FA9550-19-1-0005, ONR grant N00014-21-1-2431, and  the Sentry DHS Center of Excellence
under Award 22STESE00001-03-03.}
The authors are with ECE Department, Northeastern University, Boston, MA 02115, USA. e-mails: \{teshome.w, behzad.k, m.everett\}@northeastern.edu,
  \{camps,siami,msznaier@coe.neu.edu\}}}

\maketitle
\copyrightnotice
\thispagestyle{empty}
\pagestyle{fancy}
\fancyhf{}
\cfoot{\thepage}
\renewcommand{\headrulewidth}{0pt}
\renewcommand{\footrulewidth}{0pt}


\begin{abstract} Motivated by the problem of pursuit-evasion,
we present a motion planning framework that combines energy-based diffusion models with artificial potential fields for robust real time trajectory generation in complex environments. Our approach processes obstacle information directly from point clouds, enabling efficient planning without requiring complete geometric representations. The framework employs classifier-free guidance  training and integrates local potential fields during sampling to enhance obstacle avoidance. In dynamic scenarios, the system generates initial trajectories using the diffusion model and continuously refines them through potential field-based adaptation, demonstrating effective performance in pursuit-evasion scenarios with partial pursuer observability.
\end{abstract}

\section{Introduction} 
This paper is motivated by the problem of \MSbl{using robots} to guide crowds to safety \MSbl{in scenarios involving rapidly evolving threats, such as an active shooter or a forest fire}. This problem can be abstracted to real-time motion planning in partially known scenarios characterized by the presence of both static obstacles and dynamic,  adversarial agents.

Among the many types of motion planning algorithms, sampling-based methods~\cite{lavalle2001randomized,kavraki1996probabilistic,karaman2011sampling}  have proven effective for high-dimensional problems but often generate jerky paths requiring post-processing and suffer from computational inefficiency in sampling~\cite{petrovic2018motion}. Meanwhile, optimization-based approaches~\cite{zucker2013chomp,schulman2014motion,mukadam2018continuous, kalakrishnan2011stomp} use gradient or second-order information to efficiently find smooth trajectories. However, these methods are limited to providing locally optimal solutions\cite{orthey2023sampling}. 

Recent learning-based approaches address some of these limitations by using data-driven models for improved planning efficiency and adaptability in high-dimensional spaces and complex environments~\cite{wang2021survey}. Among these, diffusion models have emerged as particularly promising for trajectory generation, offering powerful distribution modeling capabilities~\cite{janner2022planning}. While diffusion planners
excel at capturing multimodal behaviors and handling environmental constraints, they face computational challenges during the iterative sampling process, especially in real-time applications.
In contrast, classical methods, such as artificial potential fields (APF)~\cite{khatib1986real}, excel at local obstacle avoidance but lack the global planning capabilities of learning-based approaches. 
\begin{figure}[!t]
    \centering
    \includegraphics[width=0.45\textwidth]{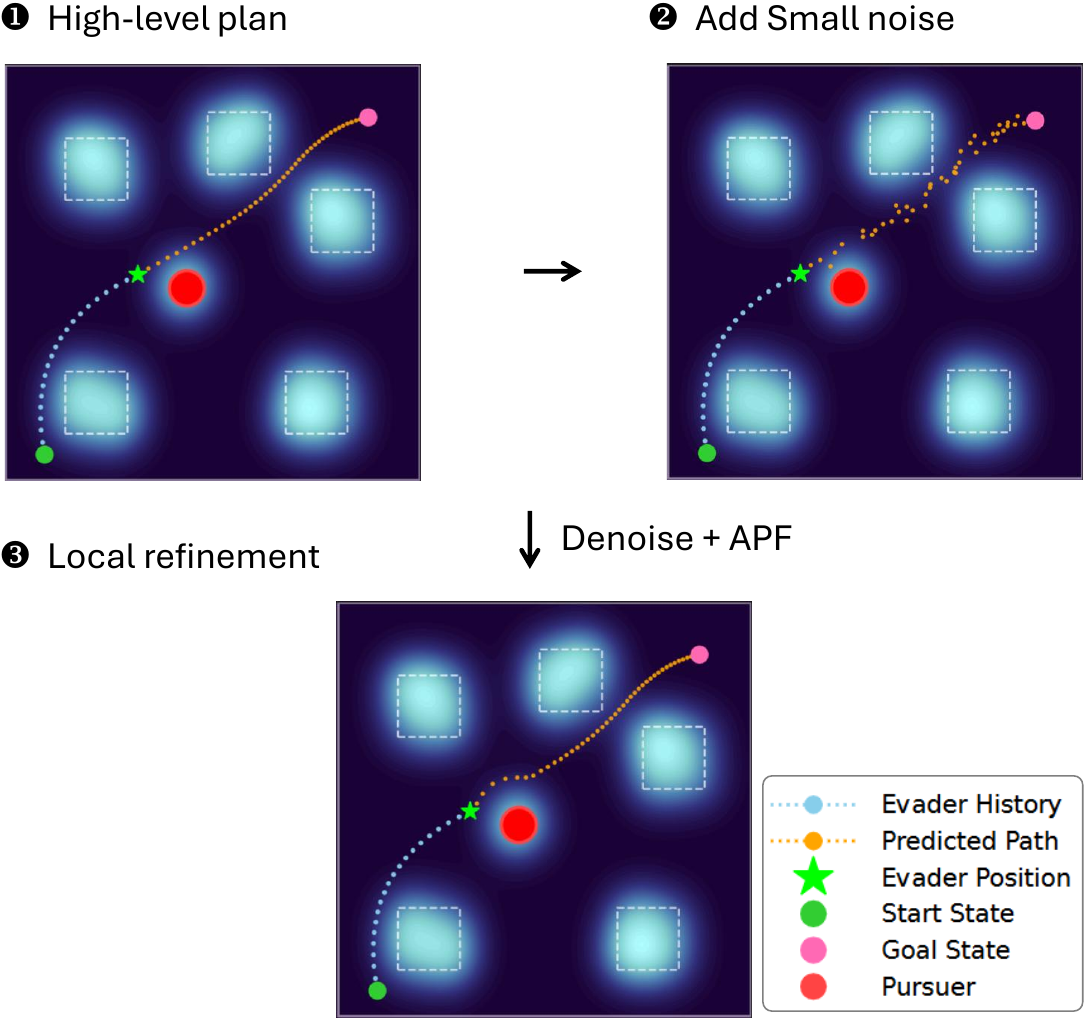}
    \caption{\small \textbf{Overview of our approach:} (1) Initial energy-based diffusion trajectory planning, where light-blue regions represent high-energy potential fields (obstacles), lightblue dots trace the evader's (green star) state history, red sphere indicates a dynamic pursuer, and dark purple areas show safe, low-energy navigation spaces. (2) When the pursuer approaches  the evader, local path exploration is performed via small noise perturbations. (3) Final trajectory refinement using denoising and APF, demonstrating adaptive path planning that avoids obstacles. }
    \label{fig:overview}
    \vskip -2em
\end{figure}
In this work, we propose a hybrid framework that integrates energy-based diffusion models with APFs, leveraging the global planning capabilities of learned models while maintaining the reactive benefits of classical methods. Our approach, outlined in Fig. \ref{fig:overview}, processes obstacle information directly from point clouds, bypassing reliance on geometric maps. This addresses a practical constraint in robotic systems, where the map may be unknown but a depth camera or LiDAR sensor could provide point cloud measurements. 
To handle  dynamic environments, we propose a hierarchical framework that integrates high level planning with real-time denoising and APF-based refinement (Figure~\ref{fig:unet-cross-attention}).  
This real-time denoising step amounts to running a few steps in a  diffusion, allowing for balancing real-time computational cost versus performance. As shown in the paper, this approach proves to be
 particularly effective in pursuit-evasion scenarios. Specifically, our contributions include:
 \begin{itemize}
     \item A real-time motion planning algorithm for environments with both static obstacles and adversarial dynamic agent, which extends energy-based diffusion models to leverage artificial potential fields and a point cloud encoder, and
     \item Demonstrations of the proposed algorithms in previously unseen scenarios, compositional scenarios, and a hardware implementation of pursuit-evasion.
 \end{itemize}

\section{Related Work}

Recent learning-based approaches have shown promise in addressing the limitations of classical methods. 
For example, MPNet~\cite{qureshi2019motion} pioneered the use of neural networks for motion planning, while \cite{strudel2021learning} improved obstacle representation using PointNet~\cite{qi2017pointnet}.
 Diffusion models enable trajectory generation through iterative denoising. Diffuser \cite{janner2022planning} introduced trajectory generation  conditioned on rewards and constraints. Motion Planning Diffusion (MPD) \cite{carvalho2023motion} extended this by learning diffusion models as priors for trajectory distributions. Most closely related to our work is Potential-Based Diffusion Planning \cite{luo2024potential}, which learns separate potential functions that can be composed at test time. While \cite{luo2024potential} requires explicit knowledge of obstacle locations, our method differs by processing point cloud representations and integrating artificial potential fields during sampling.

In dynamic environments, methods like RRTX \cite{otte2016rrtx} enable rapid replanning, while SIPP \cite{phillips2011sipp} improves efficiency by grouping safe configurations over time. Classical approaches such as artificial potential fields (APF) have been enhanced for smoother navigation in unstructured settings \cite{zhai2024local} and adapted for pursuit-evasion tasks \cite{tang2021pe}. Deep reinforcement learning has also been employed for dynamic obstacle avoidance \cite{hart2024enhanced, kastner2021arena}, with architectures like a two-stream Q-network for spatial-temporal planning \cite{wang2018learning}, and cooperative frameworks for multi-robot coordination \cite{han2020cooperative}.

In pursuit-evasion scenarios, \cite{zhu2018learning} developed evasion strategies using Deep Q-Networks, while \cite{qu2023pursuit} proposed multi-agent DRL systems with adversarial-evolutionary training. Additionally, \cite{zhang2022game} approached the pursuit-evasion problem from the pursuer perspective, developing a multi-UAV pursuit framework with target prediction capabilities to enhance coordination in obstacle-rich environments.

Despite these advances, DRL-based approaches generally suffer from limitations including poor generalization to novel scenarios and susceptibility to overfitting \cite{grando2024improving}. 

Our approach addresses these limitations by integrating learned global planning with reactive local planning through a hybrid diffusion-APF framework. This combines point cloud-based obstacle encoding with energy-based diffusion models, enabling both compositionality between different static obstacle sets and enhanced real-time performance for dynamic scenarios through reactive potential fields.

\section{Preliminaries}
\subsection{Problem Formulation}
In this paper, we address the pursuit-evasion motion planning problem in an environment where an evader must navigate to a goal while avoiding both static obstacles and a dynamic pursuer.
Let $\mathcal{C}$ be the configuration space of the evader, with obstacle space $\mathcal{C}_{obs}$ and free space $\mathcal{C}_{free}=\mathcal{C}\setminus \mathcal{C}_{obs}$. The environment contains static obstacles $\mathcal{O}_{static}=\{O_1, O_2, ..., O_n\}$ represented as point clouds, and a pursuer $\mathcal{P}$ with state $p(t)$ at time $t$, modeled as a sphere with radius $r_{\mathcal{P}}$. Given an initial state $s_s\in \mathcal{C}_{free}$, a goal state $s_g\in\mathcal{C}_{free}$, and the environment representation, our objective is to find a trajectory $\tau:[0,T]\rightarrow\mathcal{C}_{free}$ that satisfies the following constraints: 
\begin{itemize}
    \item Boundary conditions: $\tau(0)=s_s,\tau(T)=s_g$
    \item Remains collision-free with static obstacles: $\tau(t) \in \mathcal{C}_{\text{free}} \; \forall t \in [0,T]$, and
    \item Maintains safe distance from the pursuer: $\|{\tau(t) - p(t)}\| > r_{\text{safe}} \; \forall t \in [0,T]$.
\end{itemize}
The pursuer is only partially observable, detected when within radius $r_{detect}$ of the evader. The planner must adapt in real-time to both static obstacles and the pursuer.
\subsection{Diffusion Models}

Diffusion models   learn to generate data with the same distribution as the training data \cite{sohl2015deep,ho2020denoising}. To this effect, during the training phase, noise is added to the data through a forward diffusion process that  transforms  trajectories ${\tau}_0$ with an unknown distribution $q_o({\tau}_0)$  into Gaussian distributed ones via
$$q({\tau}_t|{\tau}_{t-1}) = \mathcal{N}({\tau}_t;\sqrt{1-\beta_t}{\tau}_{t-1},\beta_t\mathbf{I})$$ where $\beta_t$ represents the noise level at each step  $t\in \{1,\dots,N\}$ and $N$ denotes the number of diffusion steps. The Gaussian nature of this process allows for obtaining ${\tau}_N$ in a single step through  
$$q({\tau}_N|{\tau}_0) = \mathcal{N}({\tau}_N;\sqrt{\bar{\alpha}_N}{\tau}_0,(1-\bar{\alpha}_N)\mathbf{I})$$ with $\alpha_N=1-\beta_N$ and $\bar{\alpha}_N = \prod_{s=1}^N \alpha_s$. This is followed by a 
 reverse diffusion process that learns to recover the data by approximating 
$$p_\theta({\tau}_{t-1}|{\tau}_t) = \mathcal{N}({\tau}_{t-1};\mu_\theta({\tau}_t,t),\sigma^2_t \mathbf{I})$$ where $\mu_\theta$ is the learnable mean function and $\sigma^2_t \mathbf{I}$ is the covariance with variance parameter $\sigma^2_t=\beta_t(1-\bar{\alpha}_{t-1})/(1-\bar{\alpha}_t)$. This can be achieved by training a neural network $\epsilon_\theta$ with parameters $\theta$ to predict the noise component by minimizing $$\mathcal{L}(\theta) = \mathbb{E}_{{\tau}_0,t,\epsilon \sim \mathcal{N}(0,\mathbf{I})} \|\epsilon - \epsilon_\theta({\tau}_t,t)\|^2$$ where ${\tau}_t = \sqrt{\bar{\alpha}_t}{\tau}_0 + \sqrt{1-\bar{\alpha}_t}\epsilon$. For conditional diffusion models, additional inputs $\mathbf{c}$ are included, modifying the noise prediction network to $\epsilon_\theta({\tau}_t,t,\mathbf{c})$. 
Once the model is trained, data with a distribution $q_o(\tau)$ can be generated by applying the learned denoising process to random noise with distribution $\mathcal{N}({\tau}_N;\sqrt{\bar{\alpha}_N}{\tau}_0,(1-\bar{\alpha}_N)\mathbf{I})$.

\begin{figure*}[h!]
    \vspace{2mm}
    \centering
    \includegraphics[width=.865\textwidth,trim=0.cm .cm 0.cm .cm,clip]{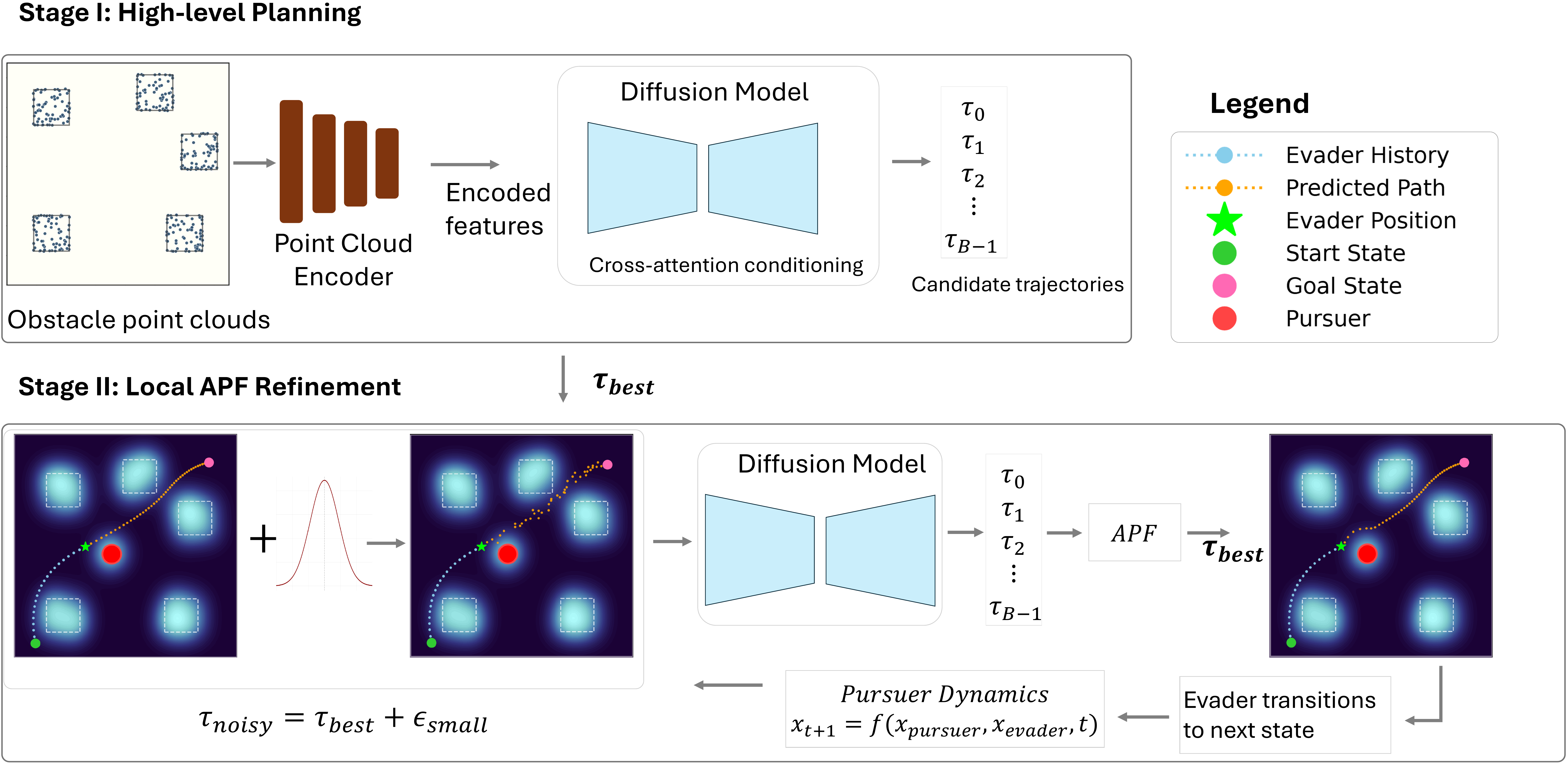}
\caption{\small \textbf{Trajectory planning pipeline for a Pursuit-Evasion case:} In the high-level planner, obstacle point clouds are processed by a pre-trained encoder to obtain a low-dimensional encoding of the point cloud features. The diffusion model then generates multiple candidate trajectories conditioned on start/goal states and these obstacle features. The best trajectory is passed to the low-level planner, which performs refinement when a pursuer approaches the evader. This refinement combines denoising with APF while conditioning on the evader's state history. The refined trajectory is then used as input for the next planning iteration, creating an adaptive system that continuously responds to the pursuer's movements while navigating through the environment.}
    \label{fig:unet-cross-attention}
\end{figure*}
\subsection{Energy-Based Diffusion Models}
Recent work has shown that diffusion models can benefit from energy-based parameterization as an alternative to the traditional score-based approach \cite{du2023reduce}. Score-based models \cite{song2020score} directly estimate the score function $-\nabla_\tau \log p_\theta(\tau_t)$. Energy-based diffusion models take a different approach by first defining a parameterized  energy function 
$f_\theta(\tau, t)=-\|s_\theta(\tau, t)\|^2$, where $s_\theta(\tau, t)$ represents a neural network with vector output \cite{du2023reduce}.  The corresponding score is given by:  
\[
    \epsilon_\theta(\tau, t) = -\nabla_\tau f_\theta(\tau, t) =  \nabla_\tau \|s_\theta(\tau, t)\|^2 \]
While this approach requires computing an additional backward pass gradient, it provides significant benefits by enabling the composition of multiple diffusion models through their unnormalized log-probabilities~\cite{du2023reduce}.
For example, \cite{luo2024potential} used this compositional property to combine multiple motion constraints in trajectory planning.

\subsection{Artificial Potential Fields}
Artificial potential field methods \cite{khatib1986real} provide a motion planning framework by modeling the robot's configuration space as a virtual force field. In this framework, the total potential field $U(\tau) = U_{att}(\tau) + U_{rep}(\tau)$ combines attractive forces toward goals and repulsive forces away from obstacles. The robot's trajectory $\tau$ typically evolves according to $\Delta \tau=-\nabla U(\tau)$, guiding the robot toward regions of lower potential energy.
While traditional APF approaches are known to suffer from local minima issues \cite{yun1997wall, zhu2006robot}, our integration with learning-based trajectory planning offers key advantages. Specifically, this approach allows the learned model to first generate trajectory proposals, after which the APF guides these initial trajectories away from obstacles with minimal interference to the overall trajectory distribution. In our implementation, we employ a simple yet effective exponential repulsive potential field adapted from \cite{gazi2007aggregation} around obstacles to ensure collision avoidance during trajectory generation, as detailed in Section \ref{sec:planning}.
\section{Proposed Framework for Adaptive Motion Planning}\label{sec:planning}
We propose a hierarchical trajectory planning framework that combines energy-based diffusion models with artificial potential fields for both static and dynamic environments. Our method integrates high-level trajectory generation with continuous refinement to handle pursuit-evasion scenarios.

\subsection{Training Conditional Energy-Based Diffusion Model}
The overall framework is shown in Fig. \ref{fig:unet-cross-attention}. At the high-level planning stage, a conditional energy-based diffusion model learns trajectory distributions conditioned on obstacle point clouds. Each   obstacle in the scene is represented by a set of points (2D or 3D) sampled from its surface and interior. The point clouds are processed using an encoder that extracts point features, which are then processed through set transformer blocks\cite{lee2019set} to produce a multi-scale scene representation.
The scene encoding conditions the diffused trajectory features through cross-attention layers at multiple resolutions in the temporal U-Net\cite{janner2022planning}, allowing the model to incorporate obstacle information.
The energy-based training objective follows \cite{du2023reduce,luo2024potential}:
\begin{equation}
L_\text{MSE} = \|\epsilon - \nabla_{\tau} E_\theta(\tau_t(\tau_0,\epsilon),t,z)\|^2
\end{equation}
where $E_\theta$ is the energy function parameterized by a temporal U-Net architecture with cross-attention mechanisms that allow diffusion features to attend to obstacle point cloud embeddings at multiple scales and $z$ are the obstacle features. 
In this paper we adopt  classifier-free guidance (CFG)  training \cite{ho2022classifier}, where we randomly drop obstacle points. This  enables the model to jointly learn the  distribution of the trajectories, conditioned on the obstacle locations, and the unconditional motion prior. 
\begin{figure}[t] 
\vspace{-3mm} 
\begin{algorithm}[H]
\caption{Energy-Based Diffusion Planning with APF for Static Obstacles} 
\small
\begin{algorithmic}[1]
\State \textbf{Input:} initial start state $s_s$, goal state $s_g$, pretrained diffusion energy function $E_\theta$ with integrated encoder, diffusion steps $N$, static obstacle point cloud $\mathcal{O}_\text{static}$, guidance scale $w$, batch size $B$, APF parameters $\gamma$, $d_\text{thresh}$, $N_{apf}$
\State $z_\text{latent} \gets f_\text{encoder}(\mathcal{O}_\text{static})$ 
\State $\tau_N \sim \mathcal{N}(0, \mathbf{I})$
\For{$t = N, \ldots, 1$}
    \State $e_\text{cond} \gets \nabla_\tau E_\theta(\tau_t, t, z_\text{latent})$
    \State $e_\text{uncond} \gets \nabla_\tau E_\theta(\tau_t, t, \mathbf{0})$
    \State $e_\text{comb} \gets (1+w)e_\text{cond} - w e_\text{uncond}$
    \State $\mu_t \gets \frac{1}{\sqrt{\alpha_t}} \left(\tau_t - \frac{1-\alpha_t}{\sqrt{1-\bar{\alpha}_t}}e_\text{comb}\right)$ 
\If{$t < N_\text{apf}$} \Comment{\textcolor{gray}{Apply APF in final denoising stages}}
    \State $\mathbf{O}_\text{nearest} \gets \text{NP}(\mu_t, \mathcal{O}_\text{static})$ \Comment{\textcolor{gray}{Find nearest obstacle points using KD-Tree}}
    \State $\mathbf{d} \gets \mu_t - \mathbf{O}_\text{nearest}$ \Comment{\textcolor{gray}{Vector to nearest obstacles}}
    \State $d \gets \|\mathbf{d}\|$ \Comment{\textcolor{gray}{Scalar distances}}
    \State $\mathbf{F} \gets \gamma \exp(-d/d_\text{thresh}) \cdot \mathbf{d}/d$ \Comment{\textcolor{gray}{Repulsive force}}
    \State $\mu_t \gets \mu_t + \mathbf{F}$ \Comment{\textcolor{gray}{guide trajectories towards regions of lower potential energy}}
\EndIf
    \State $\tau_{t-1} \gets \mu_t + \sigma_t \xi, \quad \xi \sim \mathcal{N}(0,I)$
    \State $\tau_{t-1}[0] \gets s_s, \tau_{t-1}[H-1] \gets s_g$ \Comment{\textcolor{gray}{Apply hard conditioning}}
\EndFor
\State \Return $\tau_0$
\end{algorithmic}
\label{alg:static_planning_apf}
\end{algorithm}
\end{figure}
\subsection{Trajectory Generation with Integrated Potentials} 
For scenarios with static obstacles, we propose the trajectory generation approach detailed in Algorithm \ref{alg:static_planning_apf}. This approach  combines learned energy potentials with APF during the sampling process. It  consists of two key components: (1) a reverse diffusion process that generates trajectories using classifier-free guidance, and (2) an integration of APF during the denoising process. Given a static obstacle point cloud $\mathcal{O}_\text{static}$, we first get its latent representation through an obstacle encoder that is trained jointly with the energy function $E_{\theta}$. 
During the reverse diffusion process, we iteratively denoise the trajectory while incorporating both learned and external potentials. At each step $t$, we generate trajectories  conditioned on the start state, goal state and  obstacle features through the following denoising process \cite{ho2020denoising}:  
\begin{equation}
\tau_{t-1} = \mu_t + \sigma_t\xi, \quad \xi \sim \mathcal{N}(0,I)
\end{equation}
where $\tau_{t-1}$ represents the trajectory at timestep $t-1$ and $\xi$ is random Gaussian noise added during the reverse process if $t>1$, otherwise it is $0$.
Following \cite{ho2020denoising}, the denoised prediction $\mu_t$ is computed using: 
\begin{equation}
\mu_t = \frac{1}{\sqrt{\alpha_t}}\left(\tau_t - \frac{1-\alpha_t}{\sqrt{1-\bar{\alpha}_t}}e_\text{comb}\right)
\end{equation} 
where $e_\text{comb}$ combines the conditional and unconditional energy gradients as \cite{ho2022classifier}:
\begin{equation}
    e_\text{comb} = (1+w)e_\text{cond} - w\ e_\text{uncond}
\end{equation}
where $w$ is the guidance scale that controls the conditioning strength, $e_\text{cond}$ is the conditional energy gradient, and $e_\text{uncond}$ is the unconditional energy gradient.  The external APF is integrated with the learned potential during the final denoising stages $(t<N_\text{apf})$. The APF generates repulsive forces, which guide the trajectory towards regions of low potential, defined as:
\begin{equation}
   \mathbf{F}(\mathbf{x}, \mathcal{O}) = \gamma \exp\left(-\frac{d}{d_\text{thresh}}\right) \frac{\mathbf{d}}{d}
   \label{eq:apf}
\end{equation}
where $\mathbf{d} = \mathbf{x} - \text{NP}(\mathbf{x}, \mathcal{O})$ and $d = \|\mathbf{d}\|$. Here, $\text{NP}(\mathbf{x}, \mathcal{O})$ finds the nearest obstacle point, with parameters $\gamma$ and $d_\text{thresh}$ controlling avoidance strength and influence range, respectively. This potential field generates a repulsive force that increases exponentially as the distance to the nearest obstacle decreases. The complete trajectory update incorporating both the denoised prediction and APF becomes:
\begin{equation}
\tau_{t-1} = \mu_t + \mathbf{F}(\mu_t, \mathcal{O}) + \sigma_t\xi
\end{equation}

\textbf{Compositional Sampling:} For handling multiple sets of obstacle configurations, we leverage the compositional property of energy-based diffusion models \cite{du2023reduce,luo2024potential}. When sampling trajectories that must respect multiple obstacle configurations, we modify the energy gradient combination in Algorithm \ref{alg:static_planning_apf} as:
\begin{equation}
e_\text{comb} = e_\text{uncond} + \sum_{i=1}^{N_\text{obs}} w_i(e_\text{cond}^i - e_\text{uncond})
\end{equation}
where $e_\text{cond}^i$ represents the conditional energy gradient for the $i$-th set of obstacles, and $w_i$ is the corresponding guidance scale.
This approach enables handling previously unseen scenarios, as long as they can be decomposed into ones where models are available, by simply composing the component's energy gradients during the sampling process.
\begin{figure}[t] 
\vspace{-3.75mm} 
\begin{algorithm}[H]
\caption{Dynamic Trajectory Refinement with Pursuit-Evasion}
\small
\begin{algorithmic}[1]
\State \textbf{Input:} initial trajectory $\tau_\text{high}$ from Algorithm \ref{alg:static_planning_apf}, goal state $s_g$, dynamic simulation steps $N_\text{dyn}$, static obstacles $\mathcal{O}_\text{static}$, batch size $B$, APF parameters $\gamma$, $d_\text{thresh}$, detection range $r_\text{detect}$
\State Initialize pursuer state $\mathbf{p}_0$, executed history $\mathcal{H} \gets {\tau_\text{high}[0]}$
\For{$t = 0, \ldots, N_\text{dyn}-1$}
    \State Update pursuer: $\mathbf{p}_t \gets f_\text{dynamic}(t, \tau_\text{high}[0])$ 
    \State $\tau_M \gets \tau_\text{high} + \sigma\xi, \quad \xi \sim \mathcal{N}(0,I)$ \Comment{\textcolor{gray}{Add small noise}}
    \State $\tau_M \gets$ repeat($\tau_M$, $B$) \Comment{\textcolor{gray}{Create batch of candidates}}
    \For{$j = M, \ldots, 1$} \Comment{\textcolor{gray}{Refinement steps}}
        \State Refine $\tau_j$ using steps 5-8 from Algorithm \ref{alg:static_planning_apf}
        \If{$j < M_\text{threshold}$} \Comment{\textcolor{gray}{Apply APF in final steps}}
            \State $d_\text{pursuer} \gets \|\tau_j[t] - \mathbf{p}_t\|$
            \If{$d_\text{pursuer} < r_\text{detect}$}
                \State Apply APF for both $\mathcal{O}_\text{static}$ and $\mathbf{p}_t$ 
            \Else
                \State Apply APF for $\mathcal{O}_\text{static}$ only
            \EndIf
        \EndIf
    \EndFor
    \State $\tau_\text{smooth} \gets \text{Algorithm \ref{alg:smooth}}(\tau_0[t], \tau_0[t+n], \Delta t, n, v_\text{max})$ \Comment{\textcolor{gray}{Apply smooth transitions }}    
    \State $\tau_\text{best} \gets \text{select\_best}(\tau_{smooth})$ \Comment{\textcolor{gray}{Based on cost metric}}
    \State Update history: $\mathcal{H} \gets \mathcal{H} \cup \{\tau_\text{best}[t+1]\}$\Comment{\textcolor{gray}{Update executed state}}
   \State $\tau_\text{best}[0:|\mathcal{H}|] \gets \mathcal{H}, \tau_\text{best}[H-1] \gets s_g$ \Comment{\textcolor{gray}{Apply history and goal conditioning}}
    \State $\tau_\text{high} \gets \tau_\text{best}$
    
    \If{$\|\tau_\text{high}[t] - s_g\| < \varepsilon$}
        \State \textbf{break}
    \EndIf
\EndFor
\State \Return $\tau_\text{high}$
\end{algorithmic}
\label{alg:dynamic_refinement}
\end{algorithm}
\end{figure}
\subsection{Adaptive Trajectory Refinement for Dynamic Environments including Adversarial Agent}

In principle, this paper's motivating problem of pursuit-evasion (i.e., reaching a goal while avoiding a dynamic pursuer) can be solved through a model predictive approach: At each time instant, exploit the compositional property of Algorithm \ref{alg:static_planning_apf} to generate a trajectory conditioned on the present and predicted future positions of the pursuer, given by
\begin{equation}
   \mathbf{p}_{t+1} = \mathbf{p}_t + \Delta t \cdot v \cdot \mathbf{d}_\text{pursuit}
\end{equation}
Here $\mathbf{d}_\text{pursuit} = k_p \cdot \frac{\mathbf{s}_t - \mathbf{p}_t}{\|\mathbf{s}_t - \mathbf{p}_t\|}$, where $\mathbf{p}_t$ and $\mathbf{s}_t$ are pursuer and evader positions respectively, $v$ is the pursuer velocity, and $k_p$ controls the pursuit strength.
However, this approach is relatively slow and thus may not be able to handle rapidly changing scenarios. To address this challenge, we propose the approach outlined in Algorithm \ref{alg:dynamic_refinement}.
Dynamic adaptation is handled through continuous refinement, where the evader only considers pursuer avoidance within a specified detection radius $r_\text{detect}$. As the iteration progresses, the evader's history (executed) states are stored and used as  conditioning states for subsequent trajectory sampling. Intuitively, this approach amounts to running just a few denoising steps, starting from the current trajectory, as opposed to a complete reverse diffusion.
To maintain dynamically feasible trajectories, we implement velocity-constrained smoothing at transition points between executed and planned states. This smoothing process, detailed in Algorithm \ref{alg:smooth}, is applied to a custom window size of the trajectory segments. Our experiments use a small window size (2 or 3). After applying this smoothing, the best trajectory is selected, and the history state is updated. 
This process continues until the evader reaches within $\varepsilon$ distance of the goal state, i.e., $\|\tau_\text{high}[t] - s_g\| < \varepsilon$.\newline

\textbf{Trajectory Selection:}
From the batch of generated trajectories, we first identify collision-free candidates through obstacle checking. Among these feasible trajectories, we select the trajectory with the lowest cost using a weighted cost metric:
\(\text{cost}(\tau) = w_l\cdot L(\tau) + w_s\cdot S(\tau)\)
where $L(\tau)$ denotes path length, $S(\tau)$ measures trajectory smoothness, and $w_l$, $w_s$ are their respective weights. 
\begin{algorithm}[H]
\caption{Smoothing with Velocity Constraints}
\small
\begin{algorithmic}[1]
\State \textbf{Input:} States $s_1$, $s_2$, timestep $\Delta t$, steps $n$, max velocity $v_{\text{max}}$
\State $\Delta\mathbf{w} \gets s_2^{pos} - s_1^{pos}$ \Comment{\textcolor{gray}{Position difference}}
\State $\mathbf{v} \gets \Delta\mathbf{w}/\|\Delta\mathbf{w}\|$ \Comment{\textcolor{gray}{Unit direction}}
\State $\mathbf{v}_{\text{tar}} \gets \Delta\mathbf{w}/(n \cdot \Delta t)$ \Comment{\textcolor{gray}{Target velocity}}
\State $\mathbf{v}_{\text{base}} \gets \begin{cases} 
\mathbf{v} \cdot v_{\text{max}} & \text{if } \|\mathbf{v}_{\text{tar}}\| > v_{\text{max}} \\
\mathbf{v}_{\text{tar}} & \text{otherwise}
\end{cases}$ \Comment{\textcolor{gray}{Clamped velocity}}
\State $\mathbf{w}_t \gets s_1^{pos} + t \cdot \mathbf{v}_{\text{base}}$ \text{for} $t \in [\Delta t, 2\Delta t, ..., n\Delta t]$ 
\State $\tau_{\text{smooth}} \gets [\mathbf{w}_t , \mathbf{v}_{\text{base}}]_{t=1}^n$ \Comment{\textcolor{gray}{Concatenate positions \& velocities}}
\State \Return $\tau_{\text{smooth}}$
\end{algorithmic}
\label{alg:smooth}
\end{algorithm}
 
\section{Experiments}
Our experimental results demonstrate that our novel integration of energy-based diffusion models with APFs achieves better performance in trajectory generation for both static and dynamic environments. We evaluate our approach in static obstacle avoidance and pursuit-evasion scenarios, evaluating different aspects of generated trajectories for each case.
\subsection{Experimental Setup}
\textbf{Dataset:} We evaluate our method in maze environments with 6 and 10 obstacles. Following \cite{carvalho2023motion}, demonstrations are generated using RRT-connect \cite{kuffner2000rrt}, B-spline interpolation, and GPMP optimization \cite{mukadam2018continuous}. The training set includes 2,000 obstacle configurations, each with 20 start-goal pairs and 25 diverse trajectories. Trajectories are of size $48 \times d$, where $d=4$ (2D) or $d=6$ (3D), with 48 timesteps. For 3D, training uses 10-obstacle environments; testing is on composed setups. Evaluation uses 100 unseen configurations with 20 start-goal pairs each. In pursuit-evasion, a single unseen environment with a dynamic pursuer is used across 5 runs, each with 100 start-goal pairs. The evader always moves faster than the pursuer. 

\textbf{Evaluation Metrics:} In static obstacle environments, performance is evaluated using five metrics: \textit{Success Rate} (percentage of runs with at least one collision-free trajectory), \textit{Collision Intensity} (proportion of colliding waypoints), \textit{Waypoint Variance} (path diversity), \textit{Path Length} (PL; trajectory efficiency), and \textit{Computation Time} (average sampling time). For pursuit-evasion, we assess the evader's Goal Success Rate, Collision Rate (including obstacles and captures), and Path Length. An overall \textit{Score} combines goal achievement and collision avoidance. Metrics marked ($\uparrow$) are better when higher, and ($\downarrow$) when lower.

\textbf{Baselines:} We compare our method to several planners. For static obstacles, we benchmark against RRTC-GPMP, BIT* \cite{gammell2020batch}, and MPD \cite{carvalho2023motion}, and include our base model without APF to assess its effect. In the pursuit-evasion task, we compare our Diff-APF planner to the SAC reinforcement learning baseline \cite{haarnoja2018soft}, along with two ablations: Diff-base and Diff-SAPF (APF for static obstacles only). Results highlight the benefits of integrating artificial potential fields for improved obstacle avoidance.

\begin{table*}[!htb] 
\vspace{2mm}
\centering
\caption{\small Comparison of different motion planning methods for static obstacle environments. Bold values indicate top two results for each metric.}
\setlength{\tabcolsep}{3pt}
\begin{tabular}{@{}lccccccc@{}}
\hline
\textbf{Dataset} & \textbf{Method} & \textbf{Success} & \textbf{Collision} & \textbf{Waypoint} & \textbf{Computation} & \textbf{Path} \\
& & \textbf{Rate(\%) $\uparrow$} & \textbf{Intensity(\%) $\downarrow$} & \textbf{Variance $\uparrow$} & \textbf{Time (s) $\downarrow$} & \textbf{Length $\downarrow$} \\
\hline
  & RRTC-GPMP & $\textbf{100.00}\pm0.00$&$\textbf{0.00}\pm0.00$ &$\textbf{1.89}\pm0.27$ &$3.4\pm0.02$ &$1.78\pm0.07$ \\
& BIT* & $99.55\pm 1.43$ & $0.52\pm 0.13$&$0.28\pm0.05$ &$\textbf{0.32}\pm 0.01$ & $\textbf{1.46}\pm 0.06$\\
Maze2D-6obs & MPD & $78.8\pm 9.06$ & $7.94\pm 2.06$&$0.26\pm0.08$ &$2.11\pm 0.03$ & $\textbf{1.58}\pm 0.07$\\
 & Diff-base(ours) &$99.80\pm1.21$ &$1.94\pm1.17$ & $1.56\pm0.31$&$\textbf{0.29}\pm0.01$ &$1.84\pm0.08$ \\
  & Diff-APF(ours) &$\textbf{100.00}\pm0.00$ &$\textbf{0.35}\pm0.19$ & $\textbf{1.75}\pm0.29$&$0.47\pm0.03$ &$1.90\pm0.09$ \\
\hline
  & RRTC-GPMP & $\textbf{100.00}\pm0.00$&$\textbf{0.00}\pm0.00$ & $\textbf{2.40}\pm0.37$& $3.43\pm0.03$&$1.87\pm0.08$ \\
& BIT* & $99.80\pm 1.21$ & $\textbf{0.79}\pm 0.16$&$0.50\pm0.09$ &$\textbf{0.32}\pm 0.01$ & $\textbf{1.55}\pm 0.08$\\
 Maze2D-10obs& MPD &$59.90\pm13.35$ &$11.57\pm2.25$ &$0.31\pm0.19$ &$2.11\pm0.02$ &$\textbf{1.58}\pm0.09$ \\
 & Diff-base(ours) & $92.80\pm5.84$&$7.00\pm1.63$ &$1.02\pm0.30$ &$\textbf{0.3}\pm0.01$ &$1.78\pm0.10$ \\
 & Diff-APF(ours) &  $\textbf{99.85}\pm0.85$&$2.90\pm1.07$ &$\textbf{1.94}\pm0.60$ &$0.6\pm0.05$ &$2.10\pm0.22$ \\
\hline
\end{tabular}
\label{tab: experiments}
\end{table*}
\subsection{Implementation Details}
We build on the U-Net architecture from \cite{carvalho2023motion, luo2024potential}, using residual temporal blocks. For sampling, we use DDIM\cite{song2020denoising} with 5 diffusion steps and classifier-free guidance sampling with a guidance scale of 2.0.

Obstacle data is processed with a set-based encoder for point clouds of size $N \times 64 \times D$ (64 points per obstacle, $D{=}{2,3}$). The encoder combines PointNet-style feature extraction \cite{qi2017pointnet} with set transformer blocks \cite{lee2019set}, producing a 320-dimensional latent vector used for conditioning via cross-attention. The encoder is trained end-to-end with the diffusion model.  Training uses 100 diffusion steps, batch size 128, learning rate 1e-4, on an NVIDIA RTX 2080 Ti. Conditional dropout (p=0.2) is applied during training to improve generalization.

\subsection{Results and Discussion}
We evaluate our method in both static and dynamic obstacle scenarios. As shown in Table~\ref{tab: experiments}, in static environments, our APF-enhanced model achieves superior performance across multiple metrics, with a 100\% success rate and only 0.35\% collision intensity in 6-obstacle settings—matching RRTC-GPMP in quality but with much faster sampling (0.47s vs 3.4s). It also shows high trajectory diversity (waypoint variance: 1.75). While BIT* yields shorter paths, our method offers better success rates.

The advantages of our approach become more pronounced in complex environments with 10-obstacles. our APF model maintains a 99.85\% success rate, outperforming MPD (59.90\%) and closely matching RRTC-GPMP and BIT*. APF integration reduces collisions from 7.00\% to 2.90\% and increases waypoint variance from 1.02 to 1.94 over the base model.

\begin{figure}[H]
    \centering
    \includegraphics[width=0.45\textwidth]
   {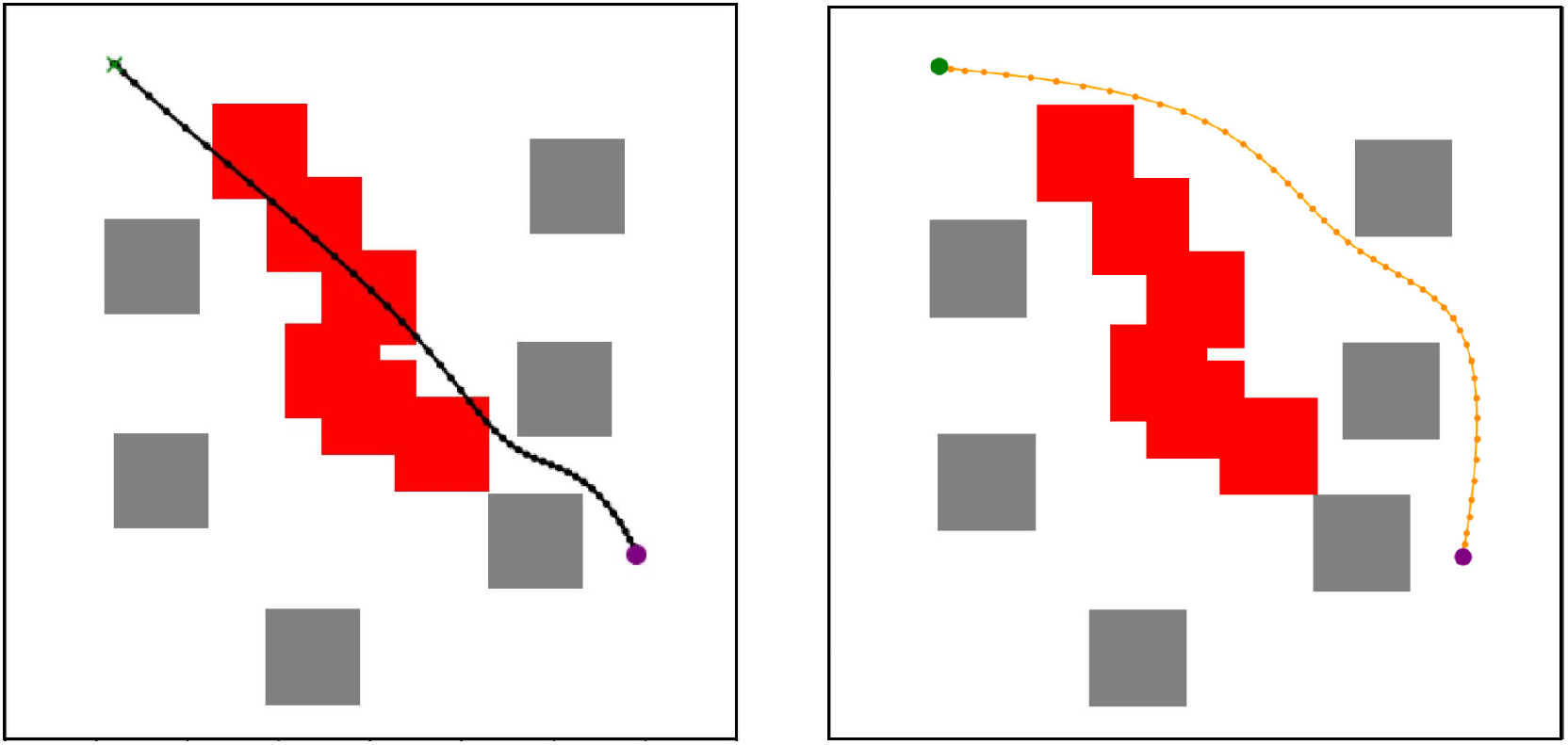}
    \caption{\small Trajectory generation comparison on Maze2D in the presence of unseen objects. \textit{left:} MDP fails to avoid the obstacles. \textit{right:} Our method successfully avoids them. }
    \label{fig:compare}
\end{figure}

For pursuit-evasion, Diff-APF achieves the highest score (84.2\%, Table \ref{tab: dynamic_table}), outperforming SAC while generating shorter, safer paths. Qualitative results (Fig. \ref{fig:compare}) show our method adapting better to unseen obstacles, generating smooth, collision-free trajectories.

\begin{figure}[!htb]
    \centering
    \includegraphics[width=0.4\textwidth, trim={0 0 0 25}, clip]{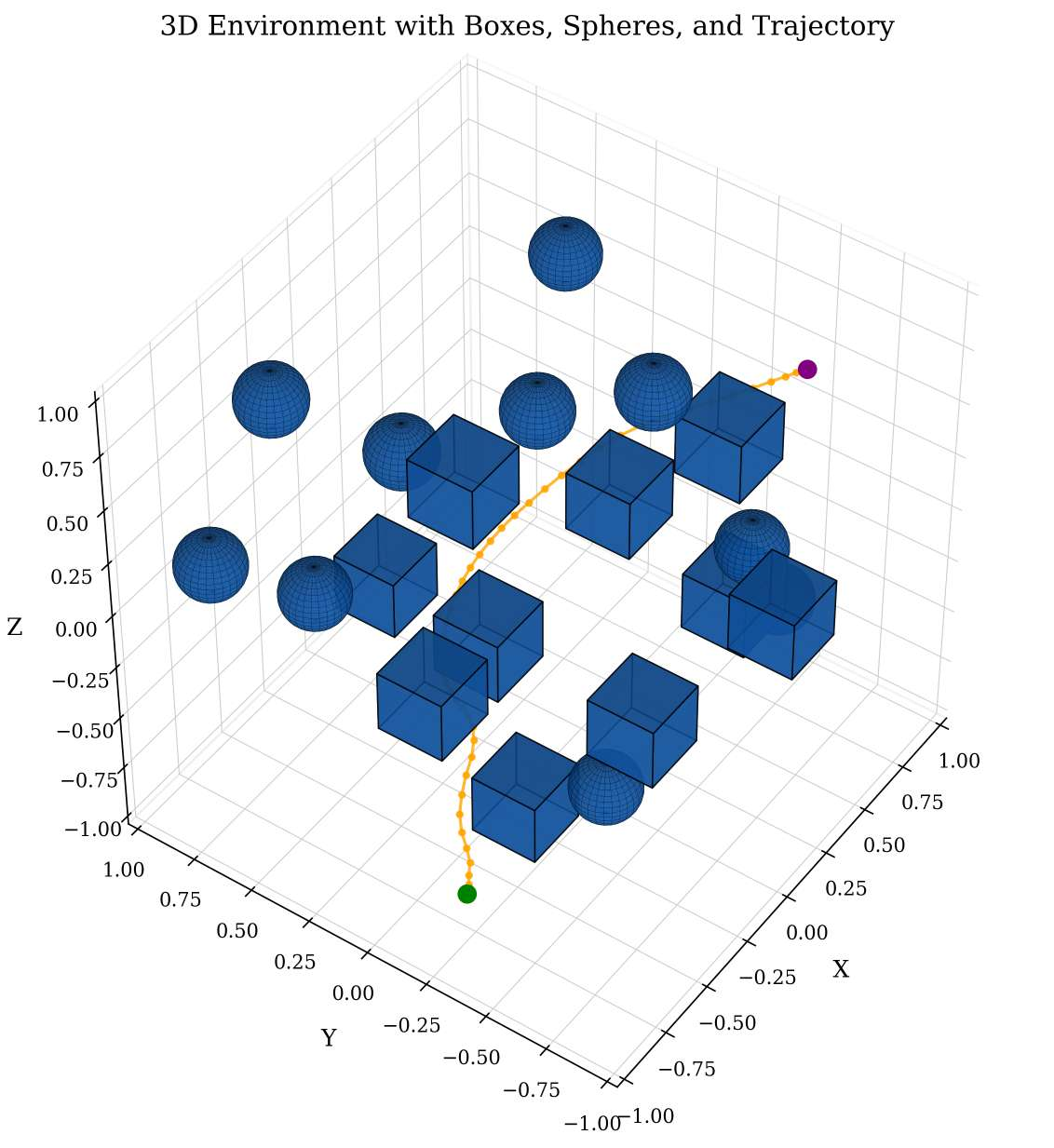}
\caption{\small Trajectory generation on Maze3D environment with box and sphere obstacles.}
    \label{fig:3d}

\end{figure}

\begin{figure*}[htbp]
    \centering
    \begin{minipage}{\textwidth}
        \centering
        \includegraphics[width=1.0\textwidth]{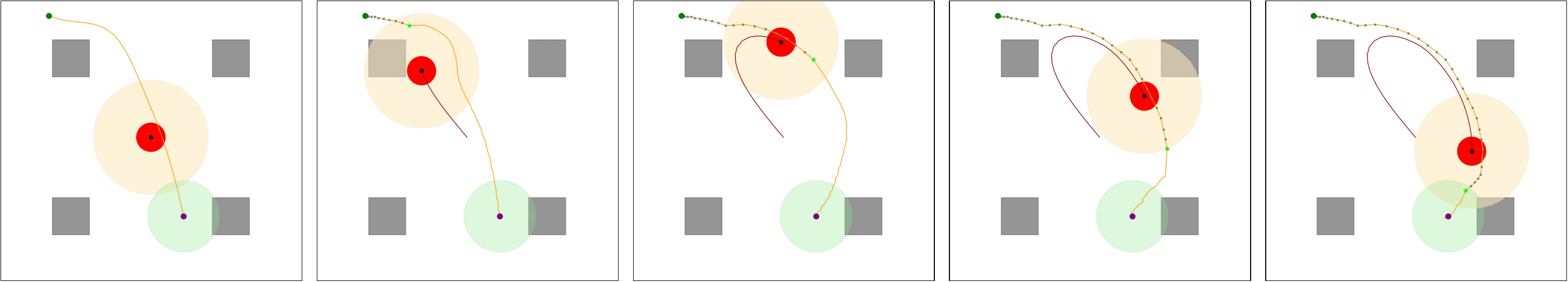}
        \small (a) \textbf{Our method} 
    \end{minipage}

    \vspace{0.1cm}
    
    \begin{minipage}{\textwidth}
        \centering
        \includegraphics[width=1.0\textwidth]{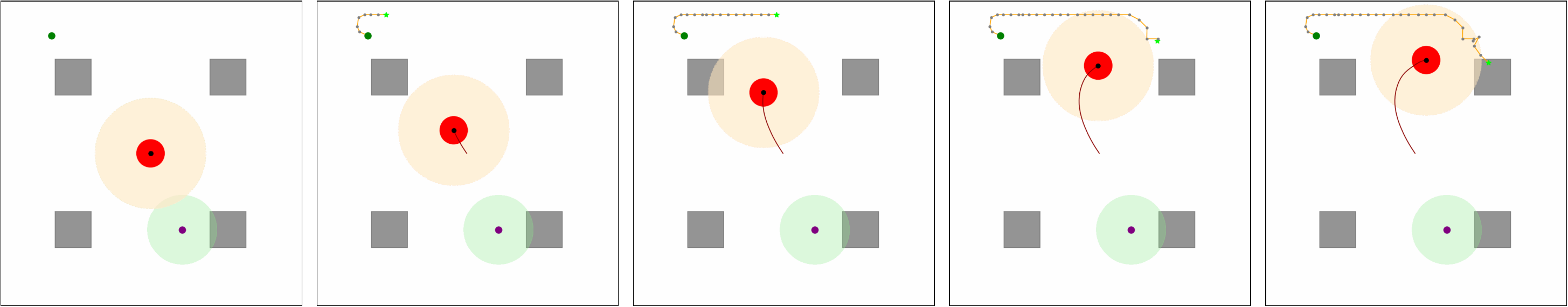}
        \small (b) \textbf{SAC method}
    \end{minipage}
    
    \caption{\small Pursuit-evasion simulation: Our method (top) successfully adapts to unseen obstacle configurations while the DRL-based SAC method (bottom) struggles. Orange lines show evader path, red circles represent pursuers with detection zones (light yellow circular regions), green areas indicate safe zones, and green/purple dots mark start/goal states.}

    \vspace{-1em}
    \label{fig:method_comparison}
\end{figure*}

In 3D (Fig. \ref{fig:3d}), our approach successfully navigates a complex arrangement of 20 static obstacles created by composing two separate obstacle configurations, demonstrating generalization. In pursuit-evasion (Fig. \ref{fig:method_comparison}), our method avoids both static and dynamic threats, unlike SAC, which struggles in unseen configurations.

\begin{figure}[htbp]
    \centering
    \includegraphics[width=0.5\textwidth]{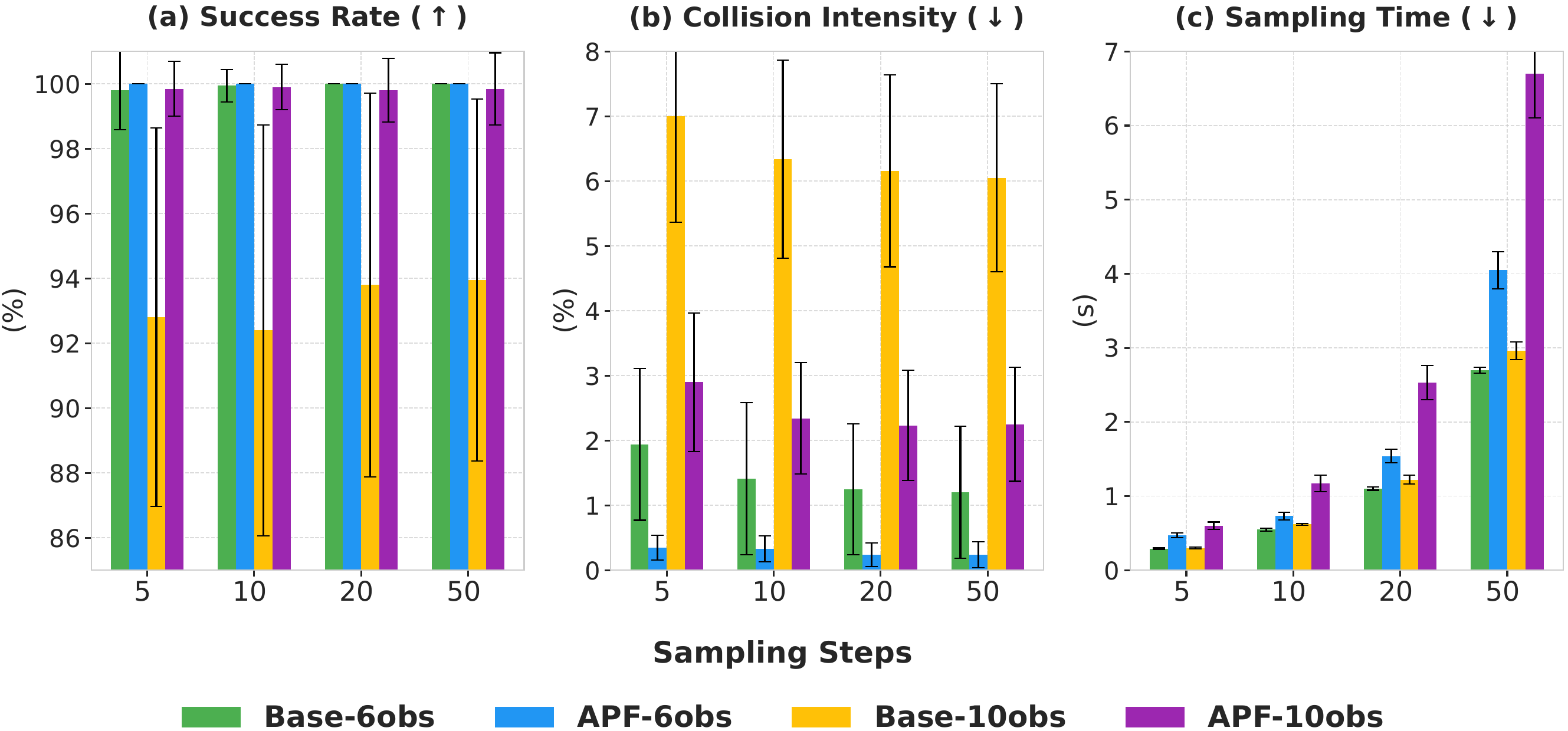}
\caption{\small Ablation on DDIM Sampling Steps: Comparison between Base and APF methods. (a) trajectory planning success rate, (b) collision intensity, and (c) computational sampling time in Maze2D environments with 6 and 10 obstacles. Error bars indicate standard deviation.}
    \label{fig:ablation_ddim}
\end{figure}

Ablation studies (Fig. \ref{fig:ablation_ddim}) show APF-enhanced models maintain >99.8\% success in complex settings. Collision intensity stabilizes after 20 DDIM steps, suggesting that increasing sampling steps further mainly impacts computational cost.  Sampling time scales linearly with the number of steps, increasing approximately 9x from 5 to 50 steps.

\begin{table}[htbp]
\centering
\caption{\small Quantitative results on Pursuit-Evasion task comparing different methods.}
\setlength{\tabcolsep}{3pt}
\begin{tabular}{lcccc}
\hline
Method & Score(\%) $\uparrow$   & PL $\downarrow$  \\
\hline
SAC  & 70.0 & 3.20 \\
Diff-base  & $30.6 $ & $1.54$ \\
Diff-SAPF & $57.6$ & $1.65$ \\
Diff-APF & $84.2 $ & $1.69$ \\
\hline
\label{tab: dynamic_table}
\end{tabular}
\end{table}

\subsection{Physical Demonstrations: Pursuer-Evader Case} 

\begin{figure}[h!]
\centering
\includegraphics[width=0.47\textwidth]{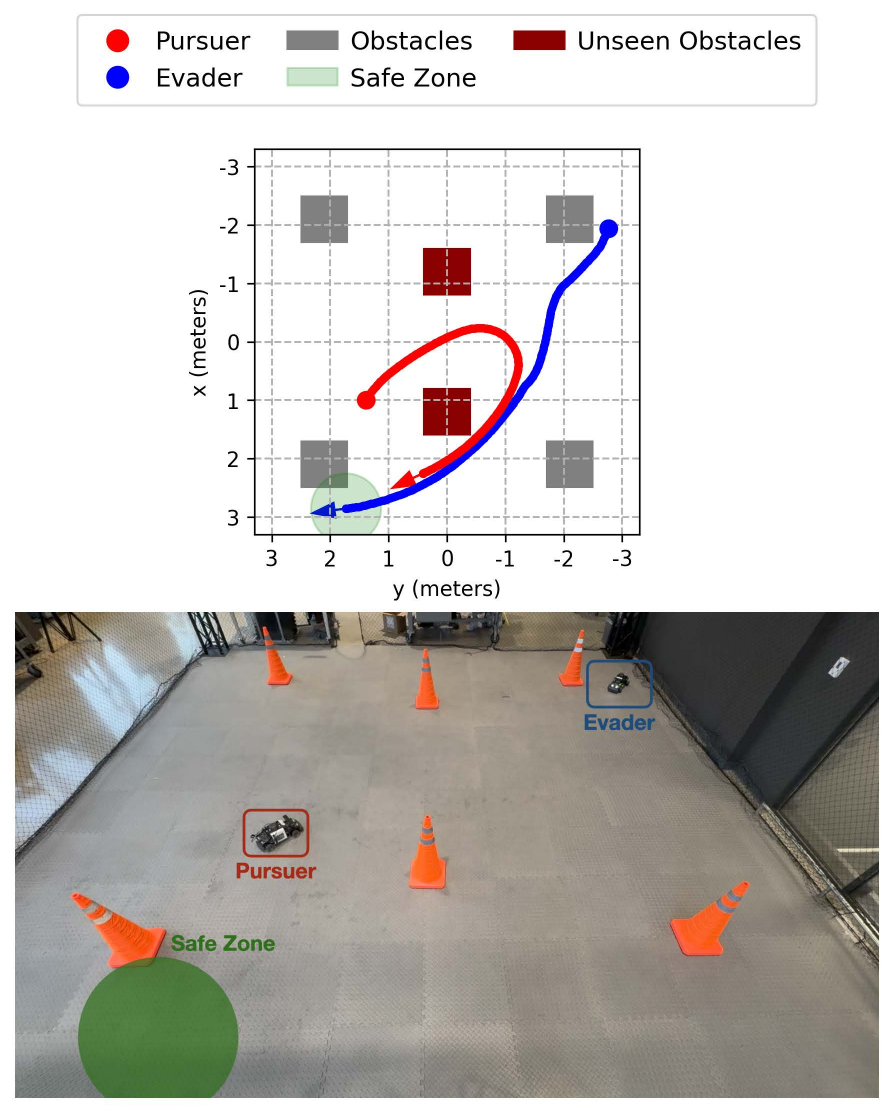}
\caption{\small Pursuit-evasion experiment showing pursuer and evader trajectory visualization (top) and corresponding lab setup (bottom).}
    \label{fig:experiments_robot}
\end{figure}

To validate our theory in an actual setup, we conducted experiments using sensor-rich RC QCars. The QCar1 by Quanser is a 1/10-scale autonomous vehicle platform designed for robotics and AI research. The platform includes key onboard sensors such as an IMU, RGB-D camera, and LiDAR, and is equipped with motion capture markers for integration with external tracking systems, making it ideal for experiments in localization, control, and navigation.  
Real-time performance evaluation using NVIDIA GeForce RTX 3080 Ti connected to the QCar1 shows our method achieves 0.15 seconds per iteration compared to MPD (0.73 seconds) and BIT* (0.3 seconds). SAC provides direct action selection (0.0004 seconds per action) rather than trajectory generation.
Figure \ref{fig:experiments_robot} shows a pursuit-evasion experiment in a 6×6 m² environment with 6 total obstacles: 4 known obstacles and 2 additional unseen obstacles. The top visualization displays the executed trajectories of both the evader and pursuer, while the bottom shows the corresponding physical lab setup.
Additional experiments and video demonstrations are available at \url{https://github.com/wondmgezahu/RAMP}.

\section{Conclusion}
We presented a novel approach combining energy-based diffusion models with artificial potential fields for motion planning. By integrating local potential fields with learned diffusion models, our method effectively handles both static and dynamic obstacles while maintaining computational efficiency. The approach demonstrates superior performance in static environments with up to 100\% success rate and significantly improves pursuit-evasion scenarios, achieving 84\% success rate. These results suggest that combining learned models with classical potential fields can enhance the robustness and adaptability of motion planning systems. 
\MSbl{Building on these promising results, future work will continue to improve the scalability of the pursuit-evasion scenario as the number of obstacles increases. Furthermore, we plan to investigate strategies for learning the APF parameters to handle increasingly complex environments.}  

\bibliographystyle{IEEEtran}
\bibliography{ref}
\onecolumn

\noindent {\large \textbf{A \quad Trajectory generation on Concave Shapes}}

\vspace{0.5cm}

\begin{figure}[!h]
\centering
\includegraphics[width=0.45\textwidth]{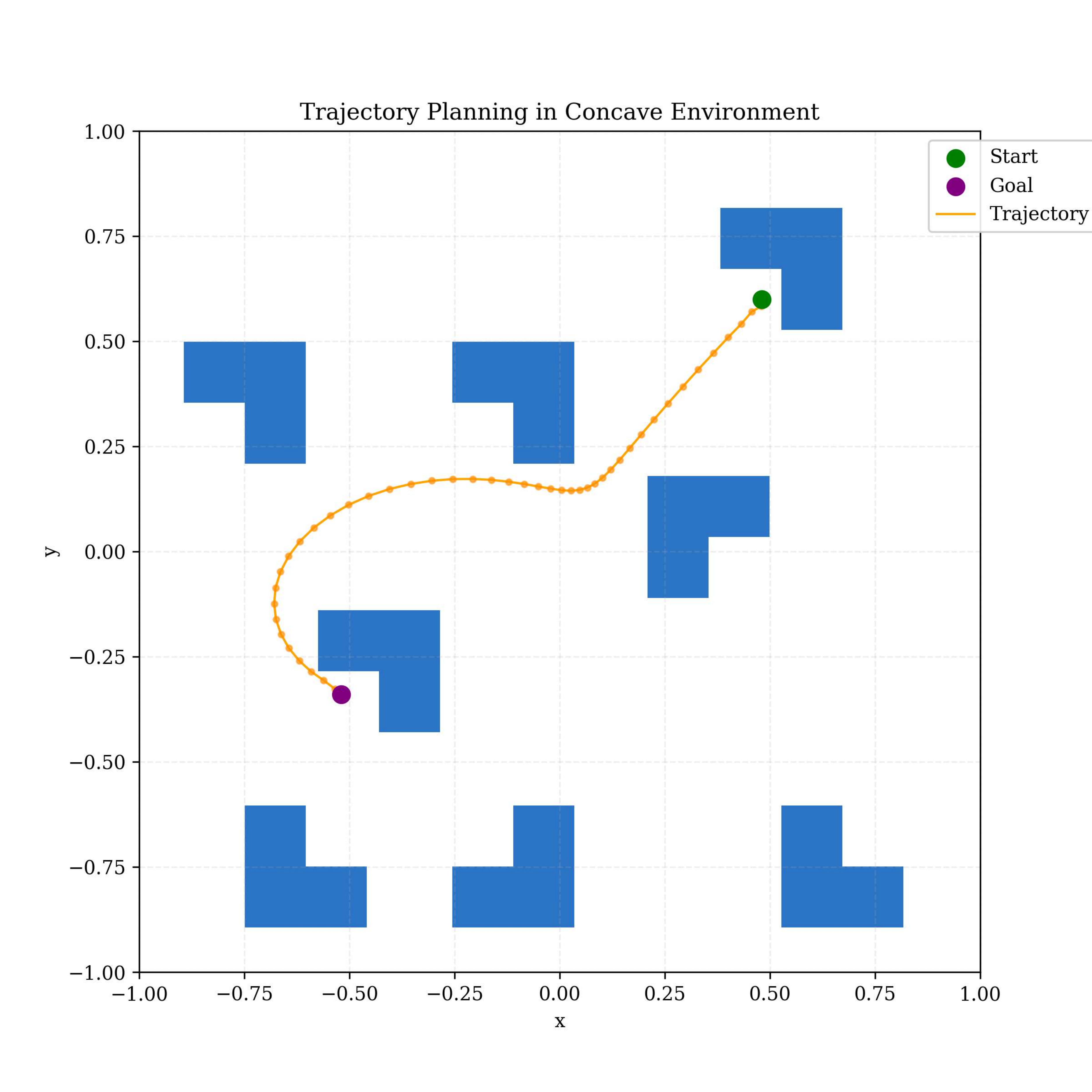}
\hspace{0.05\textwidth}
\includegraphics[width=0.45\textwidth]{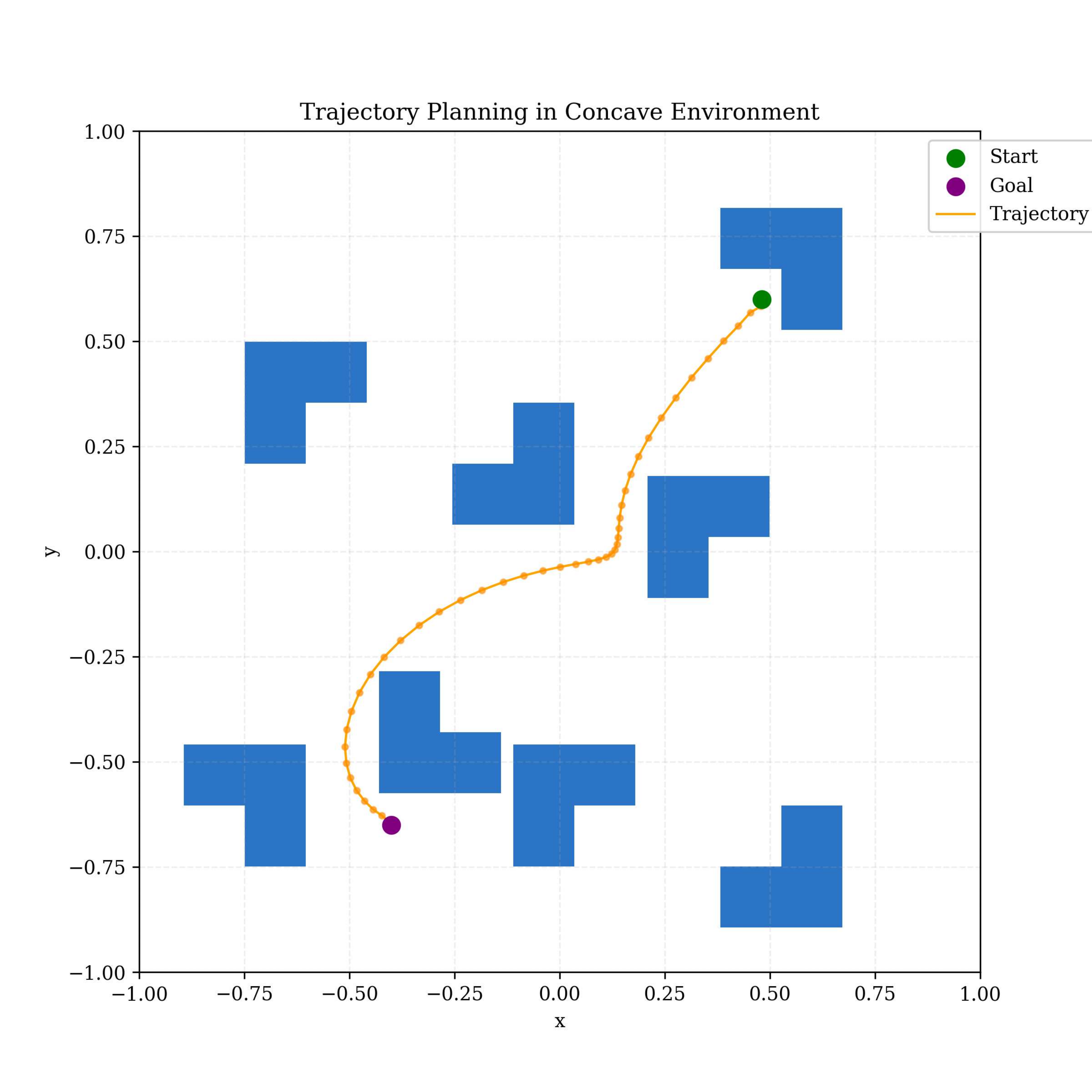}
\caption{Trajectory planning results in concave environments showing successful navigation.}
\label{fig:concave_results}
\end{figure}
\vspace{0.5cm}

\noindent {\large \textbf{B \quad Trajectory generation on Maze3D (20+) obstacles}}

\begin{figure}[!h]
    \centering
    \includegraphics[width=0.5\textwidth]{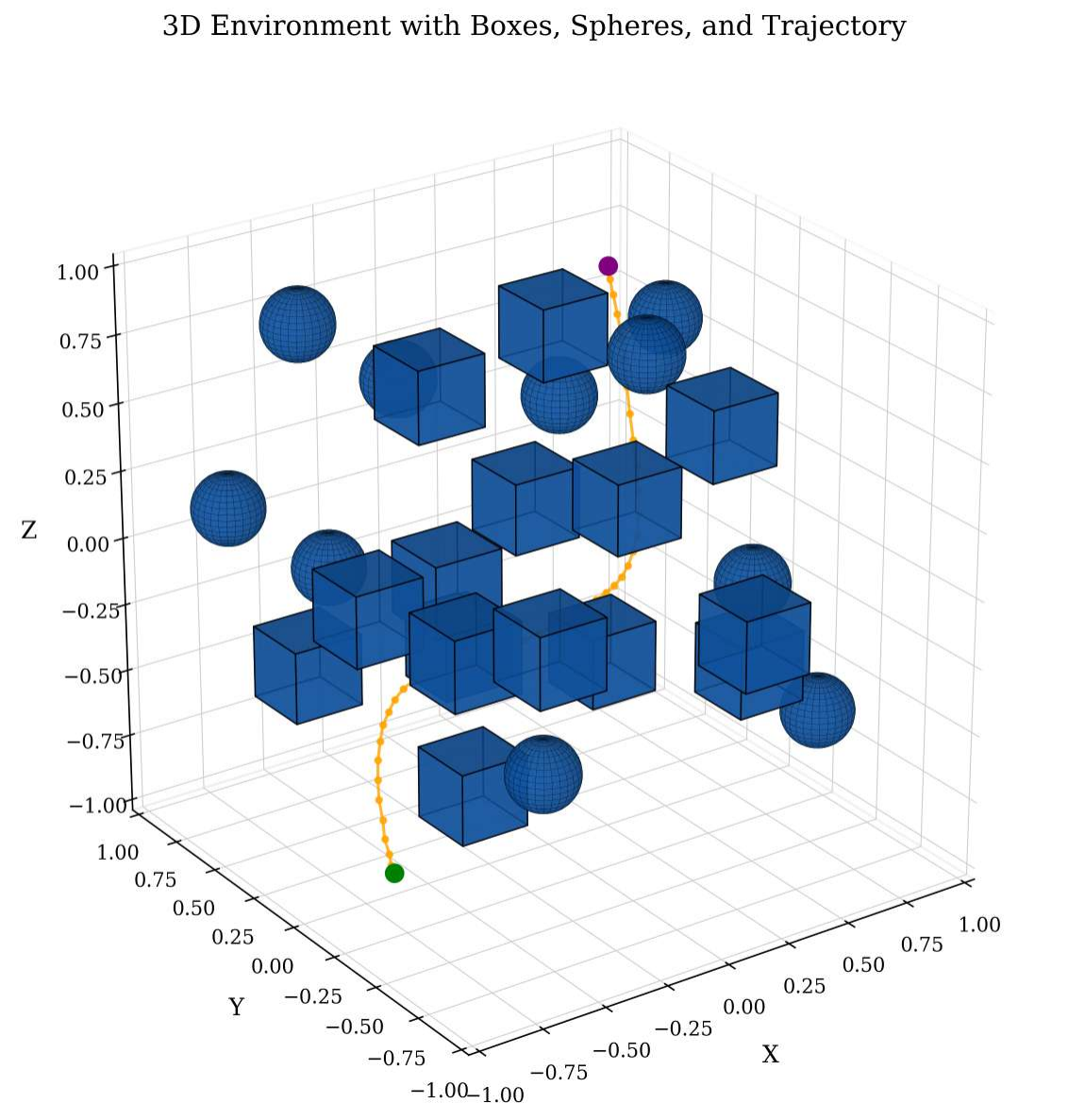}
    \caption{Trajectory planning on Maze3D environment containing 25 static obstacles. Start and goal positions are indicated by green and purple markers, respectively.}
    \label{fig:3d_25_obstacle}
\end{figure}

\vspace{0.5cm}
\end{document}